\documentclass[preprint,12pt]{elsarticle}

\usepackage{amssymb}
\usepackage{amsmath}
\usepackage{orcidlink}
\usepackage{graphicx}
\usepackage{amsmath, amssymb}
\usepackage{hyperref}
\usepackage{enumitem}
\usepackage{booktabs}
\usepackage{titlesec}
\titleformat{\paragraph}[runin]{\normalfont\bfseries}{\theparagraph}{1em}{}
\usepackage{float}
\usepackage{subcaption}
\usepackage{caption}
\usepackage{color}

\usepackage{tabularx}
\usepackage{longtable}
\usepackage[utf8]{inputenc}
\usepackage{xcolor}

\UseRawInputEncoding

\begin{document}

\begin{frontmatter}

\author{Phi-Hung Hoang}
\ead{hunghpde180523@fpt.edu.vn}

\author{Nam-Thuan Trinh}
\ead{thuantnde180305@fpt.edu.vn}

\author{Van-Manh Tran}
\ead{manhtvde180090@fpt.edu.vn}

\author{Thi-Thu-Hong Phan\corref{cor}\orcidlink{0000-0001-6880-3721}}
\ead{hongptt11@fe.edu.vn}

\cortext[cor]{Corresponding author}

\affiliation{organization={Faculty of Artificial Intelligence, \\FPT University},
            addressline={Da Nang},
            postcode={550000}, 
            country={Viet Nam}}

\title{Enhanced Fish Freshness Classification with Incremental Handcrafted Feature Fusion}

\begin{abstract}
Accurate assessment of fish freshness remains a major challenge in the food industry, with direct consequences for product quality, market value, and consumer health. Conventional sensory evaluation is inherently subjective, inconsistent, and difficult to standardize across contexts, often limited by subtle, species-dependent spoilage cues. To address these limitations, we propose a handcrafted feature--based approach that systematically extracts and incrementally fuses complementary descriptors, including color statistics, histograms across multiple color spaces, and texture features such as Local Binary Patterns (LBP) and Gray-Level Co-occurrence Matrices (GLCM), from fish eye images. Our method captures global chromatic variations from full images and localized degradations from ROI segments, fusing each independently to evaluate their effectiveness in assessing freshness. Experiments on the Freshness of the Fish Eyes (FFE) dataset demonstrate the approach's effectiveness: in a standard train-test setting, a LightGBM classifier achieved 77.56\% accuracy, a 14.35\% improvement over the previous deep learning baseline of 63.21\%. With augmented data, an Artificial Neural Network (ANN) reached 97.49\% accuracy, surpassing the prior best of 77.3\% by 20.19\%. These results demonstrate that carefully engineered, handcrafted features, when strategically processed, yield a robust, interpretable, and reliable solution for automated fish freshness assessment, providing valuable insights for practical applications in food quality monitoring.
\end{abstract}


\end{frontmatter}

\section{Introduction}

Fish is widely recognized as a highly nutritious food source, rich in protein and essential to human diets. However, ensuring the preservation and reliable evaluation of fish quality remains a significant practical challenge. Assessing fish freshness - a decisive factor for market value, product safety, and consumer health - is particularly critical. Conventional evaluation methods, which rely on human sensory judgment, are inherently subjective, error-prone, time-consuming, and difficult to reproduce consistently~\cite{Bernardo2020,Rodrigues2024}. Moreover, the visual characteristics of fish, especially the eyes, undergo subtle temporal changes influenced by species, storage conditions, and image acquisition settings. These challenges make the automatic classification of fish freshness from images a complex task that demands specialized and robust methodological approaches.

To address these issues, many studies have explored a range of techniques, from traditional machine learning with handcrafted features, which primarily exploit fine-grained differences in color, texture, and morphology to distinguish freshness levels~\cite{Lalabadi2020,Yudhana2022,Rosario2022}, to advanced deep learning models that autonomously extract discriminative representations~\cite{TaheriGaravand2020,Akgul2023,Kilicarslan2024}.
Despite such advancements, the Freshness of the Fish Eyes (FFE) dataset, an available resource for fish freshness classification, has yielded limited results, with deep learning studies reporting accuracies as low as 63.21\% using a custom MobileNetV1 architecture~\cite{Prasetyo2022} and 77.3\% with deep feature extraction combined with machine learning classifiers~\cite{Yildiz2024}. 
These limitations highlight the need for alternative approaches, particularly those leveraging handcrafted features, which remain underexplored in this context.

Inspired by the proven effectiveness of handcrafted feature extraction in other agricultural domains, such as rice seed purity classification~\cite{Phan2025Rice,Le2025Rice} and plant disease recognition~\cite{Lv2023Plant,Alsakar2024Rice,Hoang2025Potato}, we have investigated its potential on the FFE dataset. Specifically, we extract a comprehensive set of descriptors, including color-based features (color statistics, color percentiles, variance ratios, histograms) and texture-based features (LBP and GLCM), from the fish eye region. To maximize their complementary properties, we employ an incremental fusion strategy that systematically optimizes feature combinations, thereby enhancing both classification accuracy and practical applicability. The key contributions of this research are as follows:

\begin{itemize}
    \item Conducting experiments on both full images and region-of-interest (ROI) segments.
    \item Investigating a diverse set of handcrafted features, covering color and texture descriptors.
    \item Implementing an incremental fusion strategy to optimize feature combinations.
    \item Re-implementing a prior study to demonstrate the superior effectiveness of handcrafted features over deep learning approaches.
\end{itemize}

The paper is organized as follows: Section \ref{sec:related_works} reviews related work, Section \ref{sec:methodology} details the methodology, Section \ref{sec:experiment_and_setting} presents experiments, \ref{sec:result_and_discussion} provides results and discussion, and Section \ref{sec:conclusion} offers conclusions and future directions.
 
\section{Related works}
\label{sec:related_works}

Assessing fish freshness is a crucial factor in ensuring quality and safety throughout the seafood supply chain. Consequently, a substantial body of research has been devoted to automating this process. While methods based on spectroscopy and specialized chemical sensors have emerged as predominant approaches, alternative modalities such as image analysis using machine learning and deep learning have also been actively explored.

Initial research efforts primarily focused on traditional machine learning algorithms combined with handcrafted feature extraction, particularly from the characteristics of the fish's eyes and gills. Early work validated that RGB color indices from eye and gill images correlate with the decomposition process, proving especially effective after the third day~\cite{Jarmin2012}. Subsequent study, Huang et al. demonstrated that computer vision technology outperformed NIR spectroscopy for fish freshness classification, achieving 90.00\% accuracy on the prediction set compared to 80.00\% for the NIR-based model~\cite{Huang2016}. When data from both techniques were combined, the model's accuracy increased to 93.33\%. Alaimahal et al. developed a method to detect fish freshness using a watershed transform for segmentation and a wavelet transform for feature extraction~\cite{Alaimahal2017}. The classification, performed by a K-Nearest Neighbors (K-NN) classifier, achieved a correct detection rate of 90\%. More recent studies continue to demonstrate the high potential of this approach. Specifically, the efficacy of the K-Nearest Neighbors (KNN) algorithm was showcased using RGB color and GLCM texture features from fish eye images, attaining a classification accuracy of 97\%~\cite{Yudhana2022}. In another study, the combination of HSV color features with GLCM using an SVM model achieved an accuracy of 94.28\%~\cite{syarwani2022classification}. Furthermore, Ros\'{a}rio developed an automatic algorithm to detect the fish eye's pupil and extract colorfulness and brightness features, achieving 92.7\% accuracy through ensembled bagged trees~\cite{Rosario2022}.

In recent years, the field has pivoted significantly towards deep learning, particularly Convolutional Neural Networks (CNNs), which automate both feature extraction and classification. Early success was demonstrated with the VGG-16 architecture, which achieved a remarkable accuracy of 98.21\%~\cite{TaheriGaravand2020}. In the same year, Mohammadi Lalabadi et al. found that features from gills (96\%) were more effective than those from eyes (84\%)~\cite{Lalabadi2020}. Prasetyo et al. designed a customized deep learning architecture, termed MB-BE (a modified version of MobileNetV1), which achieved 63.21\% accuracy on the FFE dataset, highlighting the challenge deep learning models face in distinguishing subtle variations in fish freshness~\cite{Prasetyo2022}. Later, Akg\"{u}l et al. developed a hybrid model that successfully detected fish freshness. Their model used YOLOv5 for object detection along with Inception-ResNet-v2 and Xception for classification. This resulted in high accuracy, with the YOLOv5 and Inception-ResNet-v2 model achieving 97.67\% on anchovy and 96.00\% on horse mackerel datasets~\cite{Akgul2023}. To improve performance, recent studies have shifted toward hybrid systems that leverage pre-trained deep learning models for feature extraction and traditional machine learning methods for classification. For example, K{\i}l{\i}\c{c}arslan et al. achieved a 100\% classification accuracy by leveraging deep feature extractors (Xception, MobileNetV2) alongside classical classifiers (SVM, LR)~\cite{Kilicarslan2024}. Following this approach, Yildiz et al. achieved 77.3\% accuracy on the FFE dataset by combining VGG19 features with an ANN~\cite{Yildiz2024}, demonstrating the promise of hybrid models and the need for improved feature discrimination. The emerging trend of multi-modal feature fusion has shown promising results. As demonstrated in Balm et al., integrating image and laser reflectance data improved model robustness and achieved 88.44\% classification accuracy, outperforming single-modality approaches based on image (85.97\%) and laser (69.22\%) data~\cite{Balim2025}.

Beyond conventional RGB imaging, advanced modalities such as hyperspectral and multispectral imaging have been explored to predict chemical freshness indicators invisible to the naked eye. Khoshnoudi-Nia and Moosavi-Nasab used a multispectral system (430--1010 nm) with regression models to predict TVB-N, PPC, and sensory scores in rainbow trout, where BP-ANN performed best for PPC and GA-LS-SVM excelled for TVB-N and sensory score~\cite{Khoshnoudi-Nia2019}. Extending this line of work, Moosavi-Nasab et al. applied HSI with a Linear Deep Neural Network (LDNN) to predict TVB-N. While the LDNN showed reasonable accuracy ($R^2_P$ = 0.853; RPD = 3.001), LS-SVM remained superior ($R^2_P$ = 0.897), underscoring the competitiveness of traditional ML methods in this domain~\cite{Moosavi-Nasab2021}. Expanding further, a multi-modal spectroscopy system using VIS-NIR, SWIR, and fluorescence with LDA accurately assessed fish freshness, achieving 95\% accuracy and outperforming single-mode methods~\cite{KashaniZadeh2023}. In another approach, indirect chemical sensing methods have successfully combined pH sensors with machine learning models. Fang et al. combined red cabbage anthocyanin-based colorimetric labels with a backpropagation neural network, achieving 92.6\% accuracy in fish freshness classification~\cite{Fang2022}. Similarly, Kumaravel et al. developed a paper-based pH sensing system combined with a Random Forest model, which yielded minimal prediction errors in assessing fish freshness~\cite{Kumaravel2025}.

In summary, the literature on fish freshness assessment demonstrates a progression from handcrafted features and machine learning to sophisticated deep learning and multimodal approaches. However, as summarized in Table \ref{tab:detailed_summary_limitations_en}, many studies are constrained by small datasets, the use of a limited range of fish species, constrained environmental testing, and an over-reliance on specific modalities (e.g., images or sensors alone). These limitations underscore the need for more generalizable, efficient methods that integrate diverse features while addressing practical deployment challenges. Our work builds on this by investigating various types of handcrafted features and incremental integration to improve classification accuracy.

\begin{footnotesize}
\begin{longtable}{@{}p{0.2\textwidth} p{0.12\textwidth} p{0.2\textwidth} p{0.4\textwidth}@{}}
\caption{Summary of fish freshness assessment studies with stated limitations.}
\label{tab:detailed_summary_limitations_en}\\
\toprule

\textbf{Studies} & \textbf{Acc. (\%)} & \textbf{Methods} & \textbf{Limitations} \\
\midrule
\endfirsthead

\multicolumn{4}{c}%
{{\tablename\ \thetable{} }} \\
\toprule
\textbf{Studies} & \textbf{Acc. (\%)} & \textbf{Methods} & \textbf{Limitations} \\
\midrule
\endhead
\bottomrule
\endfoot
\bottomrule
\endlastfoot
Khoshnoudi-Nia and Moosavi-Nasab~\cite{Khoshnoudi-Nia2019} &
92.1 ($R^2_P$) &
BP-ANN &
Research was conducted on only one type of fish. A larger number of samples is needed to increase the robustness and generalizability of the models. \\
\addlinespace
Moosavi-Nasab et al.~\cite{Moosavi-Nasab2021} &
85.3 ($R^2_P$) &
HSI + LDNN &
The study was limited to rainbow trout, and more studies are required on other fish species. Validation with larger datasets is needed. \\
\addlinespace

Prasetyo et al.~\cite{Prasetyo2022} &
63.21 &
MB-BE (Modified MobileNetV1) &
The model had difficulty distinguishing between classes due to minor feature variances and insufficient features for learning.  \\
\addlinespace

Akg\"{u}l et al.~\cite{Akgul2023} &
97.67 &
YOLOv5 + Inception-ResNet-v2 &
The study's dataset was constrained to just two fish species (anchovy and horse mackerel). \\
\addlinespace

Yildiz et al.~\cite{Yildiz2024} &
77.3 &
VGG19 + ANN &
The study focused only on eye images. Future research could explore hybrid models and include other fish parts for a more comprehensive analysis. \\
\addlinespace

Balm et al.~\cite{Balim2025} &
88.44 &
Feature Fusion (ResNet50 + Laser) + MLP 
&
This study is limited by its evaluation on mackerel only and the use of a single 940 nm laser wavelength, which restricts generalizability to other species and conditions. \\
\addlinespace

Kashani Zadeh et al.~\cite{KashaniZadeh2023} &
95 &
Multi-mode Spectroscopy + LDA &
The study was limited fish species. Further research is required to validate the method's performance on other types of seafood. \\
\addlinespace

Kumaravel et al.~\cite{Kumaravel2025} &
0.0046 (MSE) &
pH sensor + Random Forest &
This method focused primarily on refrigerated conditions. The sensor's performance under different storage temperatures (e.g., frozen, room temperature) was not explored. \\
\addlinespace

\end{longtable}

\end{footnotesize}

\section{Methodology}

\label{sec:methodology}

\begin{figure}[H]
   \centering
   \hspace*{-0.8cm}
   \includegraphics[width=1.1\linewidth]{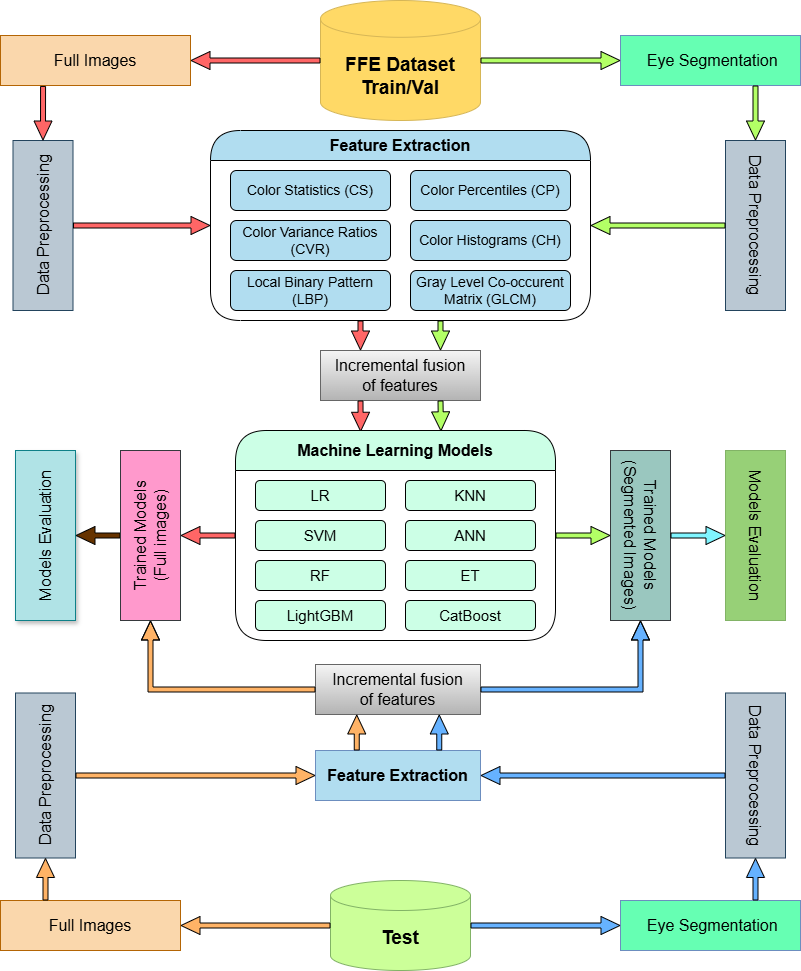}
   \caption{Overview of proposed methodology for fish freshness classification}
   \label{fig:pipeline}
\end{figure}

Figure~\ref{fig:pipeline} illustrates our proposed pipeline for fish freshness classification, which integrates image processing, handcrafted feature extraction, and machine learning. The pipeline is designed around two complementary data sources: (1) full fish images and (2) segmented eye images, isolating the fish's eye.

For feature extraction, we adopt a comprehensive approach, combining multiple handcrafted descriptors to capture subtle spoilage indicators from both sources:

\begin{itemize}
    \item Color-based features: color statistics, color percentiles, color variance ratios, and color histograms, which detect changes in brightness, hue, and overall clarity.
    \item Texture-based features: local binary patterns and gray level co-occurrence matrix, which quantify surface irregularities and the presence of cloudiness.
\end{itemize}

Our core contribution lies in a systematic, evidence-based feature fusion strategy. Instead of simply concatenating all features, we employ an \textit{incremental fusion approach} that progressively combines feature sets based on their effectiveness in prior evaluations. Individual feature groups (e.g., color-based, texture-based) and their partial combinations are first assessed, and only the most discriminative sets are integrated into the final feature representation. This ensures that the fused feature vector is both compact and optimized for classification accuracy.

In the final stage, the optimized features are used to train and evaluate a wide range of classical machine learning classifiers, including Logistic Regression (LR), k-Nearest Neighbors (KNN), Support Vector Machines (SVM), Artificial Neural Networks (ANN), Random Forests (RF), Extra Trees (ET), LightGBM, and CatBoost. Their performance is systematically compared to identify the most effective feature--classifier combinations for fish freshness prediction.

In the following sections, we present in detail the techniques employed in this study.

\subsection{Handcrafted feature extraction techniques}
\label{sec:handcrafted_feature_extraction}

This section details the extraction of handcrafted features from fish eye images. We focus on color and texture attributes that can be used to numerically represent visual characteristics and help classify different levels of freshness.

\subsubsection{Color features}

To capture diverse color characteristics and enhance robustness to lighting variations, each normalized fish eye image is transformed into multiple color spaces, including:

\begin{itemize}
    \item RGB: Provides the intensity values for the Red, Green, and Blue channels.
    \item CIELAB (Lab): Mimics human color perception by representing lightness (L*), red-green (a*), and yellow-blue (b*) components.
    \item HSV: Represents colors in terms of Hue (H), Saturation (S), and Value (V), allowing for better handling of variations in lightness and color intensity.
\end{itemize}

\textbf{a) Color statistics (CS) }

A set of statistical features is extracted from the pixel intensity values of each color channel to quantitatively characterize their distribution and complexity~\cite{Cinar2022, Mutlag2020}. These features provide a concise yet comprehensive summary of the color information within the fish eye image, capturing central tendency, variability, and the overall shape of the intensity distribution. For each color space, this procedure produces a total of 24 features (8 features per channel). The definitions of these features are summarized in Table \ref{tab:color_statistics}.

\begin{table}[H]
\centering
\footnotesize
\caption{ Features of color statistics extracted from each color channel.}
\label{tab:color_statistics}
\begin{tabular}{|l|p{9cm}|}
\hline
\textbf{Feature} & \textbf{Description} \\ \hline
Mean & The average color intensity value, indicating the overall brightness of the channel. \\ \hline
Standard Deviation & Measures the dispersion or spread of intensity values around the mean. A higher value indicates greater contrast. \\ \hline
Skewness & Indicates the asymmetry of the intensity distribution. A value of zero suggests a symmetric distribution. \\ \hline
Kurtosis & Describes the ``tailedness'' or presence of outliers in the intensity distribution compared to a normal distribution. \\ \hline
Entropy & Quantifies the randomness or unpredictability of the intensity values, which can be related to texture complexity. \\ \hline
Wavelet Moments & Statistical moments derived from wavelet decomposition, capturing texture and detail at different scales. \\ \hline
Fifth \& Sixth Moments & Higher-order moments that provide more detailed information about the shape of the intensity distribution beyond skewness and kurtosis. \\ \hline
\end{tabular}
\end{table}

\textbf{b) Color variance ratio (CVR) }

To capture the relational contrast between color channels, color variance ratios are computed by dividing the variance of each channel by that of another within the same color space. In particular, the ratios include Red-to-Green, Red-to-Blue, and Green-to-Blue in the RGB space; Hue-to-Saturation, Hue-to-Value, and Saturation-to-Value in the HSV space; and Lightness-to-a*, Lightness-to-b*, and a*-to-b* in the Lab space. By focusing on relative intensity differences between channels, these features offer additional insight into color distribution, potentially highlighting subtle patterns indicative of fish freshness. This yields nine distinct descriptors, with three ratios derived from each color space.

\textbf{c) Color percentiles (CP) }

In addition, percentile values (such as the $5^{\text{th}}$, $25^{\text{th}}$, $50^{\text{th}}$, $75^{\text{th}}$, and $95^{\text{th}}$) are computed for each component of the color spaces to capture the distribution of color intensities, providing more insight into the image's color characteristics. Fifteen features are extracted per color space (five percentiles $\times$ three components).

\textbf{d) Color histogram (CH) } 

To capture the detailed distribution of color intensities, a histogram with 16 bins per channel is computed for each individual channel in the selected color space. This bin size is chosen to balance representational detail with feature compactness, effectively reducing the occurrence of sparse bins with zero counts. For each channel, the frequency of pixels assigned to the 16 bins is calculated, and these histograms are then normalized using the L2 norm and concatenated to form the final color histogram feature vector, providing a robust and perceptually meaningful description of the eye's color profile. 

\subsubsection{Texture features}

To effectively capture surface patterns and micro-level variations on fish eyes, two complementary texture descriptors are employed: Local Binary Patterns and the Gray Level Co-occurrence Matrix. These methods are commonly used in texture analysis because they are reliable and can effectively highlight differences in texture patterns.

\textbf{Local Binary Patterns (LBP)}: The LBP operator is applied exclusively to the b* channel of the Lab color space, as this channel effectively captures color variations that are visually correlated with texture changes in fish eyes~\citep{Ojala1994}. The parameters used include 8 neighboring points (P = 8) and a radius of 1 (R = 1). The uniform LBP variant is adopted to reduce dimensionality and ensure rotational invariance. The histogram of uniform LBP codes computed from the b* channel is used as the final LBP texture descriptor.

\textbf{Gray Level Co-occurrence Matrix (GLCM)}: GLCM is a statistical method for analyzing texture based on the spatial relationship between pixel pairs. In this study, the GLCM is computed only for the b* channel in the Lab color space at a fixed pixel distance (d = 3) and four orientations ($0^{\circ}$, $45^{\circ}$, $90^{\circ}$, and $135^{\circ}$). From each GLCM, a set of Haralick features, including contrast, homogeneity, energy, and correlation, is extracted~\citep{Haralick1973}. These features are then aggregated across the four orientations using statistical measures such as the mean and range, resulting in a compact and rotation-invariant texture descriptor specific to the b* channel.

\subsection{Eye region localization and segmentation}~\label{sec:segmentation_method}

This section outlines the procedure for segmenting the eye region from fish images, forming the basis for localized feature extraction. To isolate the region of interest (ROI) and eliminate background noise, we developed an automatic segmentation algorithm based on radial intensity analysis, which detects eye boundaries without relying on deep learning models. Let $I(x,y)$ denote the original RGB input image of size $W \times H$. The overall segmentation process consists of four sequential stages:

\textbf{Step 1 (Image preprocessing):} First, the color image $I$ is converted to a grayscale intensity image $I_{\mathrm{gray}}$. To suppress high-frequency noise and stabilize the subsequent gradient calculation, a Gaussian filter $G$ with a kernel size of $7 \times 7$ is convolved with the image:
\begin{equation}
    I_{\mathrm{smooth}}(x,y) = I_{\mathrm{gray}}(x,y) \ast G_{7 \times 7}(x,y).
\end{equation}
The scanning center is defined as the geometric center of the image:
\begin{equation}
    (c_x, c_y) = \left(\frac{W}{2}, \frac{H}{2}\right).
\end{equation}

\textbf{Step 2 (Radial intensity scanning for boundary detection):} Radial scan lines are then emitted from the image center at angles $\theta \in [0, 360^\circ)$ with a step of $2^\circ$. For each direction $\theta$, the intensity profile $P_\theta(r)$ is sampled at a distance $r$ from the center:

\begin{equation}
\begin{aligned}
    x_r &= c_x + r \cos(\theta), \\
    y_r &= c_y + r \sin(\theta),
\end{aligned}
\end{equation}

where $P_\theta(r) = I_{\mathrm{smooth}}(x_r, y_r)$. To detect the eye boundary, we compute the discrete gradient of the intensity profile:

\begin{equation}
    \nabla P_\theta(r) = P_\theta(r+1) - P_\theta(r).
\end{equation}

Since the eye region is typically brighter than the surrounding skin/scales, the boundary is characterized by a significant drop in intensity (negative gradient). The candidate radius $r_\theta^{*}$ is identified at the point of the steepest negative gradient:

\begin{equation}
    r_\theta^{*} = \arg\min_{r} \, \nabla P_\theta(r),
    \quad \text{subject to } \nabla P_\theta(r) < \tau,
\end{equation}

where $\tau = -5$ is a threshold to filter out weak edges.

\textbf{Step 3 (Radius estimation and outlier removal):} Scanning across all angles yields a set of candidate radii $\mathcal{R} = \{ r_\theta^{*} \}$. To eliminate outliers caused by specular reflections, we compute the median of the set. The final radius $R_{\mathrm{final}}$ is then expanded by a factor $\lambda = 1.25$ to ensure the entire eye region is covered:

\begin{equation}
    R_{\mathrm{final}} = \left\lfloor \lambda \cdot \mathrm{median}(\mathcal{R}) \right\rfloor,
\end{equation}

\textbf{Step 4 (Mask generation and segmentation):} Finally, a binary mask $M(x, y)$ is generated using the estimated center and expanded radius:

\begin{equation}
    M(x,y) =
    \begin{cases}
        1, & \text{if } \sqrt{(x - c_x)^2 + (y - c_y)^2} \le R_{\mathrm{final}}, \\
        0, & \text{otherwise}.
    \end{cases}
\end{equation}

The segmented eye image $I_{ROI}$ is obtained by applying the mask to the original image via a bitwise AND operation:
\begin{equation}
    I_{\mathrm{ROI}}(x,y) = I(x,y) \cdot M(x,y).
\end{equation}

\begin{figure}[H]
   \centering
   \includegraphics[width=1.0\linewidth]{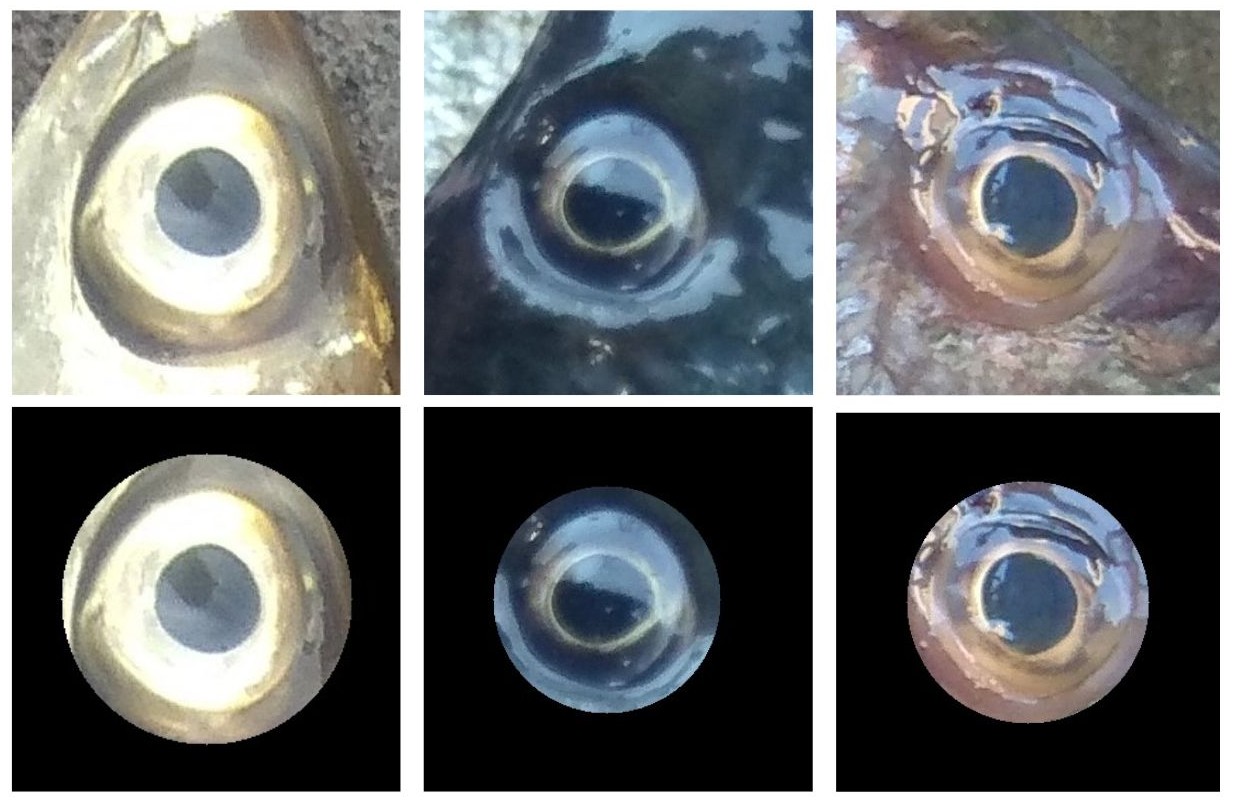}
   \caption{Comparison of eye images before and after segmentation.}
   \label{fig:segmentation_result}
\end{figure}

An example of the segmentation result is illustrated in Figure~\ref{fig:segmentation_result}.

\subsection{Incremental fusion of handcrafted features}

Handcrafted descriptors capture complementary aspects of fish-eye appearance. 
Color statistics summarize global chromatic information, variance ratios highlight inter-channel contrasts, texture descriptors (GLCM, LBP) characterize surface irregularities, and distribution-oriented features such as percentiles and histograms describe intensity variations at both local extremes and global scales. Because no single descriptor can comprehensively represent all freshness cues, combining them in a structured way is essential to obtain a richer and more reliable representation.  

To systematically integrate these complementary descriptors, we employed an incremental fusion strategy. Instead of aggregating all handcrafted features at once, descriptor groups were gradually incorporated in a stepwise manner. We began with color statistics as the baseline, then extended the representation with variance ratios to capture relational contrasts. Next, texture descriptors (GLCM and LBP) were added to encode surface irregularities 
and localized degradations. The feature space was further enriched with percentile features, which emphasize subtle intensity variations across channels, and finally complemented by color histograms, which provide a global summary of dominant chromatic patterns.  

Accordingly, seventeen feature sets (FS1-FS17) were constructed and organized into four groups:  
\vspace{0.1cm}

(i) Color statistics with variance ratios (FS1-FS5) 

\vspace{0.1cm}

- FS1: Color statistics from BGR and HSV (48 features).  

- FS2: Color statistics from BGR and Lab (48 features).  

- FS3: Color statistics from Lab and HSV (48 features). 

- FS4: Color statistics from BGR, Lab, and HSV (72 features).

- FS5: FS2 extended with Color Variance Ratios (CVR) (57 features).  
\vspace{0.1cm}

(ii) Enhanced with texture descriptors (FS6-FS7)

\vspace{0.1cm}

- FS6: FS5 + GLCM descriptors (73 features).  

- FS7: FS6 + LBP descriptors (\emph{core multimodal baseline}) (83 features).  

\vspace{0.1cm}

(iii) Percentile-augmented color features (FS8-FS14).

\vspace{0.1cm}

- FS8-FS10: FS7 + percentiles from BGR, Lab, HSV, respectively (98 features each).

- FS11-FS13: FS7 + percentiles from two spaces (BGR+HSV, BGR+Lab, Lab+HSV) (113 features each).  

- FS14: FS7 + percentiles from all three spaces (BGR, Lab, HSV) (128 features).  
\vspace{0.1cm}

(iv) Fully combined feature sets (FS15-FS17)

\vspace{0.1cm}

- FS15: FS11 + histogram from BGR (161 features).  

- FS16: FS11 + histogram from Lab (161 features).

- FS17: FS11 + histogram from HSV (161 features).  

This staged design serves two main purposes. 
First, it enables the isolation of each descriptor group, allowing us to assess its standalone discriminative capacity. 
Second, it provides a transparent framework for analyzing potential synergies, ensuring that improvements are due to complementary information rather than uncontrolled feature expansion. This incremental design not only yields richer feature representations but also provides methodological clarity, laying the groundwork for the performance evaluations presented in Section~\ref{sec:result_and_discussion}.

\subsection{Machine learning models}~\label{sec:machine_learning_models}

To address the problem of fish image classification, we implemented and evaluated the performance of a diverse range of machine learning algorithms. The selection of these models was based on their ability to handle structured data, computational efficiency, and capacity to capture both linear and complex non-linear relationships within the dataset. The following provides a detailed description of each model employed:

\textbf{Logistic Regression (LR):} LR works by fitting data to a logistic curve, which maps the linear combination of features to a probability between 0 and 1 using the sigmoid function~\cite{Hosmer2000}. The model estimates the probability of an event by transforming the linear predictor into a probability. The coefficients of the model are typically estimated using maximum likelihood estimation and indicate how much each feature affects the probability of the dependent variable.

\textbf{K-Nearest Neighbors (KNN):} K-Nearest Neighbors is a supervised, non-parametric algorithm based on the principle of proximity~\cite{Altman1991}. To classify a new data sample, KNN identifies the K closest training samples in the feature space (based on a distance metric, e.g., Euclidean). The class of the new sample is determined by a majority vote from these neighbors.

\textbf{Support Vector Machines (SVM):} Introduced by  Vapnik, SVMs are designed to determine the optimal hyperplane that best separates different classes by maximizing the margin between the hyperplane and the closest data points from each class~\cite{Vapnik1995}. To handle non-linear classification problems, SVMs utilize kernel functions to map input data into higher-dimensional feature spaces where linear separation becomes achievable. Due to their robustness in high-dimensional environments, SVMs are widely used in various classification domains, including image processing.

\textbf{Artificial Neural Network (ANN):} Artificial Neural Networks, specifically Multi-Layer Perceptrons (MLPs), are models inspired by the structure of the human brain~\cite{Rumelhart1986}. They consist of interconnected layers of neurons: an input layer, one or more hidden layers, and an output layer. Each neuron in a hidden layer applies a non-linear activation function to the weighted sum of its inputs. ANNs are capable of learning complex feature representations and non-linear relationships from data. Network architecture (number of layers, neurons per layer) and activation functions are critical design elements. The output layer employs a softmax function for multi-class classification.

\textbf{Random Forest (RF):} Proposed by Breiman, RF is an ensemble learning method that constructs multiple decision trees using bootstrap samples and random feature selection at each split~\cite{Breiman2001}. Final predictions are made by aggregating the outputs of all trees, typically through majority voting. This combination of randomness and ensembling reduces overfitting and improves classification performance.

\textbf{Extra Trees (ET):} Introduced by Geurts et al., is an ensemble method that constructs multiple decision trees and aggregates their outputs for classification or regression tasks~\cite{Geurts2006}. Unlike Random Forests, ET builds each tree using the full training data and introduces greater randomness by selecting split points at random rather than choosing the best one. This approach helps lower model variance and can improve generalization to new data.

\textbf{Light Gradient Boosting Machine (LGBM):}  is another gradient boosting framework, developed by Microsoft, focusing on speed and memory efficiency~\cite{Ke2017lightgbm}. It employs techniques such as Gradient-based One-Side Sampling (GOSS), which focuses on data instances with large gradients, and Exclusive Feature Bundling (EFB), which reduces the number of features. LightGBM grows trees leaf-wise rather than level-wise, leading to faster convergence and often better results on large datasets.

\textbf{Categorical Boosting (CB):}  is a gradient boosting algorithm developed by Dorogush et al., particularly effective at handling categorical features natively, without requiring extensive preprocessing like one-hot encoding~\cite{Dorogush2018catboost_arxiv}. It uses an innovative method called ``ordered boosting" and a target statistics-based technique for encoding categorical features to prevent target leakage and improve accuracy.

\section{Experiments}

\label{sec:experiment_and_setting}

\subsection{Data description}

\begin{figure}[H]
   \centering
   \includegraphics[width=1.0\linewidth]{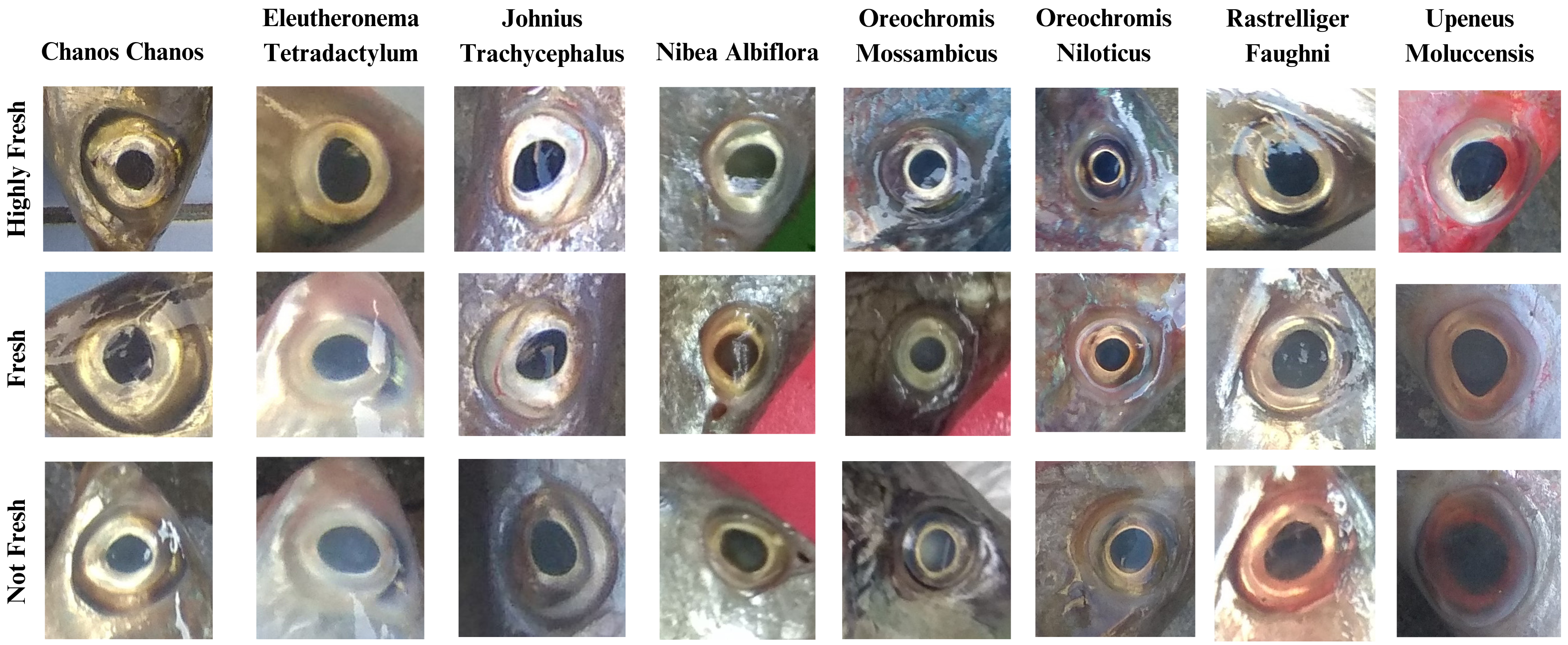}
   \caption{Sample images from the Freshness of the Fish Eyes (FFE) dataset illustrating the three freshness categories: Highly Fresh, Fresh, and Not Fresh.}
   \label{fig:dataset}
\end{figure}

The dataset employed in this study is the publicly available the Freshness of the Fish Eyes (FFE) dataset, originally introduced and described by Prasetyo et al.~\cite{Prasetyo2022}, and accessible through Mendeley data~\citep{MendeleyFFEData}. The FFE dataset consists of sRGB images of fish eyes captured to evaluate different levels of freshness. It includes images from eight fish species, with each image labeled into one of three freshness categories: `\textit{Highly fresh}' (corresponding to a 1--2 day shelf life), `\textit{Fresh}' (3--4 days), and `\textit{Not fresh}' (5--6 days). In total, the dataset comprises 4,390 images, with representative examples shown in Figure~\ref{fig:dataset}.

\begin{figure}[H]
   \centering
   \includegraphics[width=0.75\linewidth]{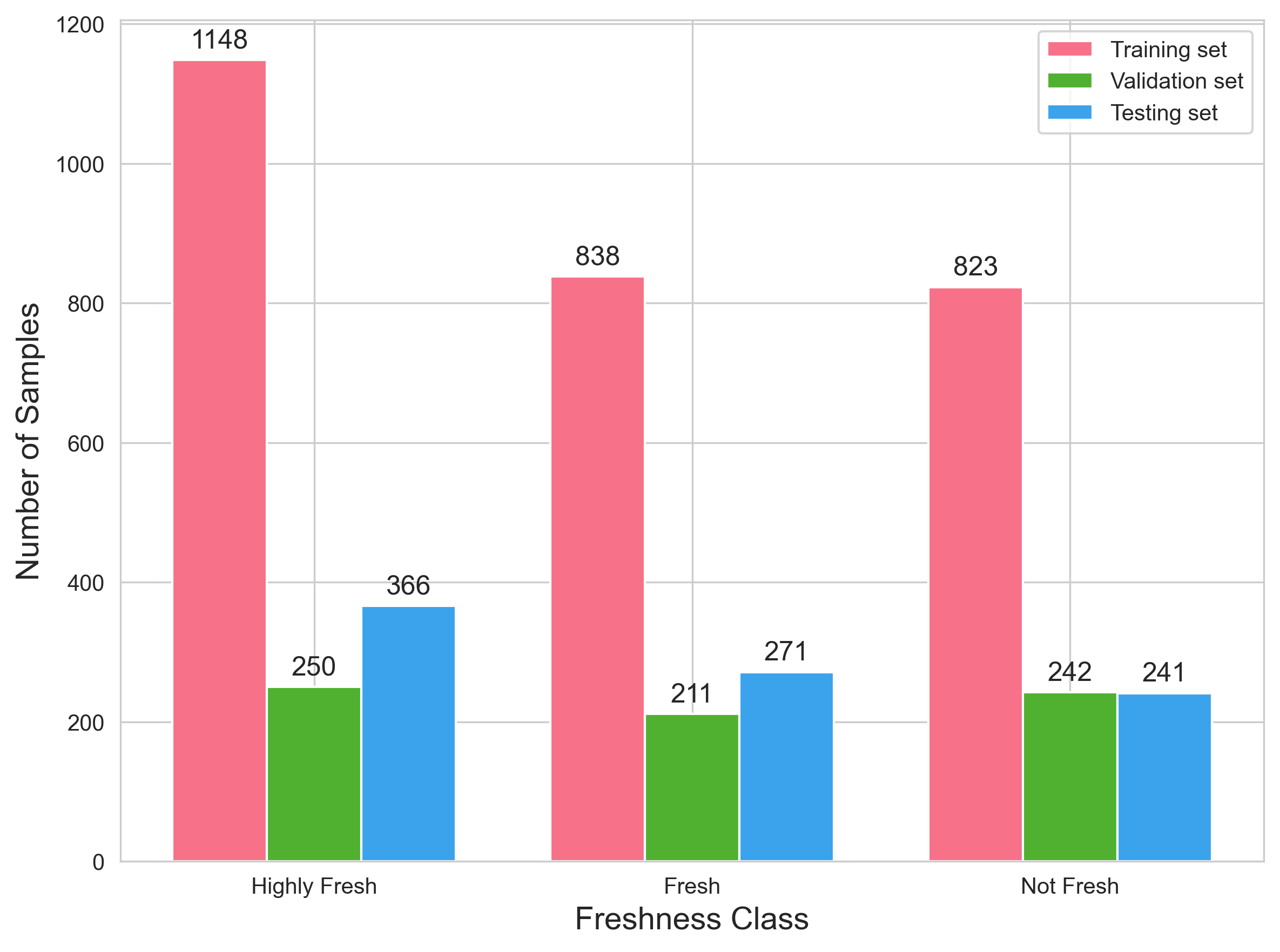}
   \caption{Distribution of the number of images in each dataset subset}
   \label{fig:train_val_test_distribution}
\end{figure}

In this study, we adopted a generalized freshness classification strategy. Specifically, instead of distinguishing among the eight fish species, we pooled all images based on their assigned freshness levels. This simplification yields a three-class problem that directly captures the visual indicators of freshness observable across species. Accordingly, the dataset used in our experiments is distributed into three categories: 1764 images labeled as highly fresh, 1320 as fresh, and 1306 as not fresh. The dataset was stratified and initially divided into 80\% for training and 20\% for testing. The training subset was then further stratified and split into 80\% for training and 20\% for validation, as illustrated in Figure~\ref{fig:train_val_test_distribution}.

\subsection{Evaluation protocols}

To ensure reliable generalization assessment and fair comparison with prior studies, two complementary evaluation protocols were adopted in this work.

\textbf{Protocol A (PA -- hold-out evaluation)}

Protocol A is used as the primary evaluation setting to assess the effectiveness of the proposed handcrafted feature extraction and incremental fusion strategy under a standard hold-out evaluation scheme. This protocol follows the stratified train--test split strategy adopted in earlier foundational studies on the FFE dataset \cite{Prasetyo2022}. Specifically, the dataset is split into training and test subsets using a stratified hold-out scheme to preserve class distributions. All models are trained and evaluated solely on the original images, ensuring that each test sample is completely unseen during training. Results obtained under Protocol A reflect the generalization performance of the proposed method and form the basis for all main analyses and conclusions in this paper.

\textbf{Protocol B (PB -- replication setting)}

Protocol B follows the experimental configuration established in previous studies \cite{Yildiz2024} to facilitate a direct comparison. In this setting, data augmentation (consisting of image flipping) is performed prior to dataset splitting, and the evaluation is conducted using 10-fold cross-validation. This protocol is specifically adopted to replicate reported results and to evaluate the proposed handcrafted feature fusion method under identical experimental conditions. Given that augmentation precedes splitting, results obtained under Protocol B should be interpreted as upper-bound performance within the original experimental framework rather than a measure of real-world generalization.

\subsection{Evaluation metrics}

In this study, the model's performance is evaluated using four commonly used classification metrics: accuracy, precision, recall, and F1-score. These metrics provide a comprehensive evaluation of the model's ability to correctly classify fish eyes into three freshness categories: highly fresh, fresh, and not fresh.

\begin{itemize}

    \item Accuracy: Measures the proportion of correctly classified instances out of the total number of instances.
    \begin{equation}
        Accuracy = \frac{TP + TN}{TP + TN + FP + FN}
    \end{equation}

    \item Precision: Indicates the proportion of true positive predictions among all instances predicted as positive.
    \begin{equation}
        Precision = \frac{TP}{TP + FP}
    \end{equation}

    \item Recall: Represents the proportion of true positive predictions among all actual positive instances.
    \begin{equation}
        Recall = \frac{TP}{TP + FN}
    \end{equation}

    \item F1-Score: The harmonic mean of Precision and Recall, providing a single score that balances both metrics.
    \begin{equation}
        F1 = 2 \cdot \frac{Precision \cdot Recall}{Precision + Recall}
    \end{equation}
    
\end{itemize}

These metrics are computed for each of the three classes and averaged to evaluate the overall performance of the model on the test set.

\subsection{Experiment setting}

All experiments were carried out on a Windows 11 system (AMD Ryzen 7 6800H CPU, 16 GB RAM, NVIDIA GeForce RTX 3050 GPU, 4 GB) with PyCharm as the main development environment. Deep learning tasks were additionally executed on Kaggle to leverage dedicated GPU resources. Classical machine learning models were implemented using scikit-learn (including gradient boosting libraries), while PyTorch was employed for deep learning components.

For machine learning experiments, hyperparameters were optimized separately via Random search on the training folds, and the best-performing configurations for the full-image pipeline are summarized in Table~\ref{tab:ml_full_image_hyperparameters}.

\begin{table}[H]
\footnotesize
\centering
\caption{Hyperparameter values for ML models yielding best performance on full images.}
\label{tab:ml_full_image_hyperparameters}
\begin{tabular}{|l|p{9.8cm}|}
\hline
\textbf{Model} & \textbf{Hyperparameter values} \\
\hline
LR & C=1.0, max\_iter=300, penalty=l2, solver=liblinear \\
\hline
KNN & algorithm=ball\_tree, n\_neighbors=17, metric=minkowski, p=2, weights=distance \\
\hline
SVM & C=100, coef0=0.01, gamma=scale, kernel=rbf \\
\hline
ANN & activation=tanh, solver=Adam, learning\_rate=adaptive, early\_stopping=False, max\_iter=500, alpha=$10^{-3}$, hidden\_layer\_sizes=(128) \\
\hline
RF & criterion=gini, max\_depth=15, n\_estimators=300 \\
\hline
ET & bootstrap=True, criterion=entropy, min\_samples\_leaf=1, min\_samples\_split=3, n\_estimators=180, max\_depth=35 \\
\hline
LGBM & boosting\_type=dart, class\_weight=balanced,  n\_estimators=300, max\_depth=50, num\_leaves=35 \\
\hline
CB & loss\_function=MultiClass, eval\_metric=Accuracy \\
\hline
\end{tabular}
\end{table}

\section{Results and discussion}
\label{sec:result_and_discussion}

This section presents and analyzes the experimental results obtained from our study. We begin with evaluations on the full set of fish-eye images to establish baseline performance. Next, we investigate the effect of segmentation on classification accuracy and provide a detailed analysis of its contribution. Finally, we compare our findings with related studies in the literature to highlight both the effectiveness and the novelty of the proposed approach.

\subsection{Results on full images}~\label{sec:results_on_full_images}

All results reported in this subsection are obtained under Protocol A (PA), following a clean hold-out evaluation without data augmentation.

\subsubsection{Performance analysis on individual feature types}

To establish a performance baseline, each handcrafted feature category was first evaluated independently on the original images. The results, summarized in Table \ref{tab:individual_result_on_original_images}, highlight the relative discriminative capacity of color, texture, and relational descriptors in fish-eye freshness classification.

Among the evaluated color spaces, CIELAB proved particularly effective, an advantage attributable to its perceptual design that decouples the lightness channel (L*) from the two chrominance channels (a* and b*). The color statistics method achieved the highest overall accuracy of 70.96\% with RF, demonstrating the utility of higher-order statistical moments in capturing perceptually salient color information. Complementary methods, such as color percentiles  and color histograms, also produced competitive accuracies of 68.56\% with ET and 67.77\% with LGBM, respectively, further reinforcing the centrality of color as a discriminative indicator of freshness.  

By contrast, descriptors designed to encode local surface texture or relational color attributes performed substantially worse when used independently. LBP reached a maximum accuracy of 55.58\% with ANN, GLCM achieved 56.04\% with LGBM, and CVR attained 61.62\% with RF. These results suggest that, although such features capture certain aspects of surface irregularities or chromatic relationships, they lack sufficient discriminative power to serve as reliable standalone predictors of freshness.  

This analysis highlights the limitations of individual descriptors. Color features effectively capture global chromatic changes but overlook fine surface details, whereas texture features encode local patterns without broader context. Their complementary nature strongly motivates the integration of multiple feature types through fusion. The following sections investigate this strategy to construct a more robust and comprehensive classification model.

\begin{table}[H]
\caption{Accuracy of ML models using individual feature type (\%).}
\footnotesize
\centering
\label{tab:individual_result_on_original_images}
\begin{tabular}{|l|c|c|c|c|c|c|c|c|}
\hline
\textbf{Feature Set}  & \textbf{LR}  & \textbf{KNN} & \textbf{SVM} & \textbf{ANN} & \textbf{RF} & \textbf{ET} &   \textbf{LGBM}   &    \textbf{CB}   \\
\hline
CS (BGR)              &     60.71    &     58.66    &     64.35    &     63.21    &    63.21    &    62.76    &        61.62      &       62.64       \\
\hline
CS (HSV)              &     59.91    &     62.30    &     66.51    &     63.67    &    66.29    &    65.60    &        67.31      &       64.69       \\
\hline
CS (Lab)              &     62.64    &     67.08    &     68.22    &     69.70    & \textbf{70.96} & \textbf{70.05} & \textbf{70.27} & \textbf{70.50}  \\
\hline
CVR                   &     53.64    &     57.86    &     56.83    &     54.67    &    61.62    &    60.36    &        59.34      &       60.02       \\
\hline
CP (BGR)              &     59.45    &     58.89    &     64.46    &     63.67    &    60.71    &    61.28    &        61.39      &       60.25       \\
\hline
CP (HSV)              &     58.77    &     60.14    &     62.53    &     61.96    &    66.40    &    65.60    &        62.98      &       64.35       \\
\hline
CP (Lab)              &     60.82    &     64.46    &     67.20    &     64.81    &    68.11    &    68.56    &        66.40      &       66.29       \\
\hline
CH (BGR)              &     57.86    &     57.63    &     61.28    &     61.96    &    63.90    &     62.07   &        64.92      &       62.87       \\
\hline
CH (HSV)              &     58.65    &     60.82    &     62.64    &     62.19    &    66.97    &     66.74   &        67.77      &       66.97       \\
\hline
CH (Lab)              &     57.29    &     58.31    &     62.64    &     61.73    &    63.44    &     65.72   &        64.46      &       63.55       \\
\hline
LBP                   &     51.14    &     50.68    &     51.78    &     55.58    &    52.16    &     51.37   &        51.14      &       49.89       \\
\hline
GLCM                  &     53.76    &     53.53    &     54.33    &     53.64    &    56.04    &     55.35   &        55.01      &       52.28       \\
\hline
\end{tabular}
\end{table}

\subsubsection{Incremental fusion analysis: Performance of combined feature types}
\textit{
{i) Color statistics with variance ratios (FS1--FS5)}}

\vspace{0.2cm}
The initial experiments on individual feature types demonstrated that color-based descriptors, particularly from the Lab space, offered the most promising results. This subsequent phase therefore examined whether combining color statistics across spaces could yield a synergistic effect, and further assessed the contribution of relational color features.

The results in Table \ref{tab:combined_color_features_results} show that Feature set 2 (BGR + Lab) achieved the strongest performance, reaching 71.98\% with LGBM. This reflects the complementary nature of the two spaces: Lab offers perceptual uniformity aligned with human vision, while BGR provides raw sensor-level information.

By contrast, Feature set 4, which combines all three spaces (BGR, Lab, HSV), did not improve performance (70.62\%). This outcome suggests that the inclusion of HSV introduced redundancy or less discriminative cues, highlighting the importance of selective rather than indiscriminate feature fusion.

The most notable gain was observed with Feature set 5, where Feature set 2 was extended with Color Variance Ratios (CVR). This configuration achieved the highest accuracy of 72.67\% (LGBM), confirming that CVR adds complementary relational information. While color statistics capture overall distributions, CVR emphasizes inter-channel contrasts that may reflect biochemical changes in fish eyes during spoilage. These findings confirm that selectively integrating complementary descriptors, rather than simply aggregating all available cues, yields more effective and discriminative feature representations.

\begin{table}[H]
\caption{Accuracy of ML models using combined color statistics features (\%).}
\footnotesize
\label{tab:combined_color_features_results}
\centering
\begin{tabular}{|l|c|c|c|c|c|c|c|c|}
\hline
Feature set   & LR & KNN & SVM & ANN & RF & ET & LGBM & CB \\
\hline
Feature set 1 & 63.67 & 64.58 & 68.56 & 68.56 & 67.43 & 67.88 & 68.34 & 68.00 \\
\hline
Feature set 2 & 63.55 & 67.08 & 69.25 & 69.25 & 70.61 & 70.39 & \textbf{71.98} & 69.59 \\
\hline
Feature set 3 & 63.67 & 67.43 & 70.05 & 69.13 & 70.62 & 69.82 & 70.38 & 69.48 \\
\hline
Feature set 4 & 64.12 & 66.86 & 70.05 & 69.48 & 70.05 & 70.05 & 70.62 & 70.50 \\
\hline
Feature set 5 & 62.87 & 66.40 & 69.70 & 68.79 & 71.30 & 71.18 & \textbf{72.67} & 71.07 \\
\hline
\end{tabular}
\end{table}

\textit{ii) Enhanced with texture descriptors (FS6--FS7)}

\vspace{0.2cm}

Building on the previously identified optimal color-based feature set (Feature set 5), this stage of the study examined whether texture descriptors could add complementary information and improve classification performance. As shown in Table \ref{tab:combined_color_features_and_texture_features_results}, we incorporated GLCM and LBP features. Importantly, instead of applying these methods to a simple grayscale image, we extracted them from the b* channel of the CIELAB color space. This channel represents the blue--yellow spectrum, which is closely associated with biological decay.

Adding GLCM features (Feature set 6) resulted in a noticeable improvement, raising the accuracy to 73.01\% with LGBM. This indicates that analyzing texture specifically in the b* channel provides useful information  about the spread and uniformity of yellowing or bluish clouding on the eye surface, both of which are subtle indicators of spoilage.

When LBP features (Feature set 7) were incorporated, the classification accuracy further increased to 75.85\%. Unlike GLCM, which summarizes broader spatial relationships, LBP captures fine-scale variations that emerge during spoilage. These include localized spots, early signs of cloudiness, and subtle edge changes around the cornea. By adding this layer of detail, LBP enhanced the feature representation and allowed the models to distinguish freshness levels more effectively. The improvement was not limited to LGBM: all classifiers (LR, KNN, SVM, ANN, RF, ET, CB) also demonstrated consistent gains.

Overall, these findings demonstrate that texture information extracted from the b* channel can meaningfully complement color statistics. By combining global descriptors such as GLCM with localized descriptors such as LBP, the feature representation becomes more comprehensive, which in turn improves the ability of machine learning models to distinguish different freshness levels.

\begin{table}[H]
\caption{Accuracy of ML models using combined color statistics and texture features (\%).}
\footnotesize
\label{tab:combined_color_features_and_texture_features_results}
\centering
\begin{tabular}{|l|c|c|c|c|c|c|c|c|}
\hline
\textbf{Feature set} & \textbf{LR} & \textbf{KNN} & \textbf{SVM} & \textbf{ANN} & \textbf{RF} & \textbf{ET} & \textbf{LGBM} & \textbf{CB} \\
\hline 
Feature set 6        &     64.35   &     66.51    &     69.70    &     70.62    &     70.50   &    71.18    &     \textbf{73.01}     &     71.98   \\
\hline
Feature set 7        &     64.69   &     67.77    &     71.07    &     71.18    &     71.75   &    71.53    &     \textbf{75.85}     &     73.80   \\
\hline
\end{tabular}
\end{table}

\textit{iii) Percentile-augmented color features (FS8--FS14)}

\vspace{0.2cm}

This phase of the investigation examined whether the strong performance of the BGR and CIELAB combination, observed earlier in the statistical analysis (Table~\ref{tab:combined_color_features_results}), would also hold when applying more granular distributional features. The underlying assumption was that if this pairing is genuinely robust, it should remain effective not only in capturing central tendencies but also in describing the extremes of the color distribution.

When each color space was evaluated separately (Feature sets 8--10), the results revealed a clear hierarchy. Percentiles from CIELAB produced a substantial improvement, raising the accuracy of LGBM to 76.31\%, whereas BGR and HSV percentiles failed to outperform the baseline. This suggests that the perceptual uniformity of CIELAB plays a decisive role in capturing subtle, nonlinear shifts in color distribution associated with spoilage, while the other spaces introduced more noise than useful information.

Building on this finding, the analysis then explored combinations of two color spaces (Feature sets 11--13). The BGR+Lab pairing proved to be particularly effective, with Feature set 12 achieving the best overall accuracy of 76.42\% under LGBM and also boosting CatBoost performance to 74.37\%. In contrast, combinations involving HSV did not yield competitive results, highlighting that BGR+Lab provides the most complementary and non-redundant source of information.

Extending the approach to include all three spaces simultaneously (Feature set 14) did not enhance performance further, with accuracy dropping to 75.63\%. The inclusion of HSV-derived features introduced redundancy and conflicting signals, confirming that simply adding more features is not an effective strategy. Instead, the results highlight that a selective, knowledge-driven approach 
focused on the complementary strengths of BGR and CIELAB provides the most reliable pathway for constructing discriminative color features in this task.

\begin{table}[H]
\footnotesize
\caption{Accuracy of ML models using percentile-augmented feature sets (\%).}
\label{tab:combined_color_features_and_color_percentile_results}
\centering
\begin{tabular}{|l|c|c|c|c|c|c|c|c|}
\hline
\textbf{Feature set} & \textbf{LR} & \textbf{KNN} & \textbf{SVM} & \textbf{ANN} & \textbf{RF} & \textbf{ET} & \textbf{LGBM} & \textbf{CB} \\
\hline 
Feature set 8        &     67.20   &     67.43    &     71.30    &     72.21    &     71.07   &    70.96    &     75.51     &     74.03   \\
\hline
Feature set 9        &     65.38   &     68.68    &     70.96    &     73.01    &     72.55   &    71.98    &     \textbf{76.31}     &     73.58   \\
\hline
Feature set 10       &     66.29   &     68.56    &     71.64    &     72.10    &     71.98   &    72.78    &     75.28     &     74.83   \\
\hline
Feature set 11       &     67.43   &     68.22    &     71.75    &     72.67    &     71.18   &    72.21    &     75.74     &     75.74   \\
\hline
Feature set 12       &     67.43   &     68.68    &     70.73    &     71.87    &     71.18   &    71.18    &     \textbf{76.42}     &     74.37   \\
\hline
Feature set 13       &     66.97   &     68.79    &     71.98    &     73.23    &     72.55   &    71.87    &     75.63     &     74.87   \\
\hline
Feature set 14       &     67.88   &     67.65    &     71.53    &     71.18    &     71.64   &    72.32    &     75.63     &     73.92   \\
\hline
\end{tabular}
\end{table}

\textit{iv) Fully combined feature sets (FS15--FS17)}

\vspace{0.2cm}

In the final stage of feature engineering, we investigated whether adding a binned distributional descriptor, specifically a color histogram, could capture remaining visual information. Our initial experiments incorporated histograms from the BGR and CIELAB spaces (Feature sets 15 and 16), both of which had previously shown strong performance with statistical and percentile-based features. 
However, as shown in Table \ref{tab:full_combined_features_results}, the results were unexpected. Neither the BGR nor the CIELAB histogram improved classification accuracy. In fact, combining CIELAB with a histogram (Feature set 16) slightly reduced performance across most models. 
This indicates that the detailed information from these color spaces had already been effectively captured by statistical moments and percentiles, making the addition of a coarse histogram largely redundant.

In contrast, the HSV histogram (Feature set 17) produced a notable improvement. 
This configuration achieved 77.56\% accuracy with LGBM and showed consistent gains across multiple classifiers. For example, the accuracy of ANN increased to 74.72\% and Extra Trees to 73.12\%, demonstrating that the added information was broadly useful rather than specific to a single model.

The success of the HSV histogram can be attributed to its ability to capture the global color profile of the image. Unlike detailed metrics, the histogram, especially of the Hue channel, summarizes dominant color trends 
that are highly indicative of spoilage. Degradation often produces a systematic shift in hue, which is clearly reflected in the histogram. This provides a complementary signal 
that earlier features could not fully capture.

These findings show that the most effective feature representation does not rely on a single color space. Optimal performance is achieved through a carefully designed fusion of multiple color spaces and feature types, 
each contributing unique and complementary information.

\begin{table}[H]
\footnotesize
\caption{Accuracy of ML models using fully combined feature sets (\%).}
\label{tab:full_combined_features_results}
\centering
\begin{tabular}{|l|c|c|c|c|c|c|c|c|}
\hline
\textbf{Feature set} & \textbf{LR} & \textbf{KNN} & \textbf{SVM} & \textbf{ANN} & \textbf{RF} & \textbf{ET} & \textbf{LGBM} & \textbf{CB} \\
\hline 
Feature set 15       &     67.43   &     68.45    &     71.18    &     72.10    &     71.53   &    72.10    &     \textbf{76.54}     &     74.60   \\
\hline
Feature set 16       &     69.36   &     67.31    &     71.75    &     72.89    &     72.89   &    69.82    &     75.63     &     74.60   \\
\hline
Feature set 17       &     67.43   &     68.45    &     72.44    &     74.72    &     72.10   &    73.12    &     \textbf{77.56}     &     73.35   \\
\hline
\end{tabular}
\end{table}

\subsection{Results on segmented images}

To investigate whether focusing exclusively on the eye-the primary region of interest-affects classification performance, we evaluated the proposed approach using segmented eye images under Protocol A, enabling a direct comparison with the corresponding results obtained from full images. By removing the surrounding background and tissues, we aimed to examine whether the models could better focus on the visual cues of freshness in the eye itself and to assess the importance of the contextual information from the surrounding regions. The results, summarized in Table \ref{tab:individual_result_on_segmented_images}, reveal that the surrounding tissues and background provide complementary cues that are critical for accurate freshness classification.

\begin{table}[H]
\footnotesize
\caption{Accuracy of ML models using individual feature type on segmented images (\%).}
\centering
\label{tab:individual_result_on_segmented_images}
\begin{tabular}{|l|c|c|c|c|c|c|c|c|}
\hline
\textbf{Feature Set}  & \textbf{LR}  & \textbf{KNN} & \textbf{SVM} & \textbf{ANN} & \textbf{RF} & \textbf{ET} &   \textbf{LGBM}   &    \textbf{CB}    \\
\hline
CS (BGR)              &     52.16    &     49.43    &     59.57    &     57.86    &    49.66    &    49.09    &        51.82      &       49.77       \\
\hline
CS (HSV)              &     52.39    &     55.24    &     58.66    &     56.78    &    55.58    &    56.59    &        55.35      &       56.15       \\
\hline
CS (Lab)              &     51.71    &     53.87    &     59.79    &     59.99    &    59.57    &    57.97    &        59.57      &       59.23       \\
\hline
CVR                   &     44.31    &     51.03    &     52.85    &     54.44    &    54.67    &    53.76    &        54.10      &       53.42       \\
\hline
CP (BGR)              &     45.22    &     45.79    &     50.34    &     50.68    &    48.18    &    47.15    &        48.41      &       46.81       \\
\hline
CP (HSV)              &     44.76    &     51.03    &     50.80    &     50.23    &    51.48    &    52.05    &        50.09      &       50.11       \\
\hline
CP (Lab)              &     45.22    &     46.01    &     50.34    &     50.23    &    48.18    &    46.92    &        48.97      &       46.81       \\
\hline
CH (BGR)              &     48.86    &     54.67    &     56.72    &     57.18    &    60.14    &     61.16   &        61.85      &       59.45       \\
\hline
CH (HSV)              &     47.95    &     57.06    &     55.92    &     59.00    &    61.96    &     62.87   &        62.53      &       61.85       \\
\hline
CH (Lab)              &     48.41    &     56.38    &     56.83    &     57.06    &    61.39    &     59.91   &        61.85      &    \textbf{63.55}     \\
\hline
LBP                   &     46.24    &     48.75    &     48.75    &     50.23    &    49.89    &     49.54   &        48.97      &       46.24       \\
\hline

GLCM                  &     47.04    &     49.09    &     50.46    &     50.00    &    51.03    &     51.25   &        50.80      &       48.06       \\
\hline
\end{tabular}
\end{table}

The most evident finding is the substantial decline in classification accuracy when using segmented images compared with full images. The best score on segmented data was only 63.55\% (color histograms in Lab space with CatBoost), which is 7.41\% lower than the peak accuracy of 70.96\% obtained from full images. This drop indicates that background regions, rather than introducing noise, provide complementary context that improves classification. Subtle cues such as the gradual color transitions between the eye and surrounding skin, or the fine texture of adjacent scales, appear to be essential for reliable freshness assessment. Removing these cues leaves the models with less informative inputs, resulting in poorer performance.

In addition, there is a noticeable shift in the relative effectiveness of feature descriptors. With full images, color statistics yielded the highest accuracies (up to 70.96\%), but on segmented images their performance fell sharply (mostly $\sim$55--60\%). By contrast, color histograms  became the most reliable feature type, consistently outperforming CS across color spaces and reaching the top score of 63.55\%. This suggests that once the eye region is isolated and becomes more uniform, the overall distribution of dominant colors captured by histograms provides a more stable signal than statistical measures such as mean or variance.

Meanwhile, texture-based features (LBP, GLCM) and relational color descriptors (CVR, CP) consistently produced near-random accuracies, often only 45--52\%. Their ineffectiveness stems from the small and relatively homogeneous surface of the isolated eye, which lacks the rich textures and inter-regional color relationships present in the full images.

Since the accuracy of each individual feature type dropped significantly on segmented images, we did not experiment with pairwise combinations as in the full-image case. Instead, we directly tested the fusion of all feature types through Feature sets 15--17 to evaluate whether comprehensive integration could recover performance (Table \ref{tab:combined_color_features_results_on_segmented_images}). The results show only moderate improvements: the best score was 68.11\% (Feature set 16 with LGBM), which is still lower than the top accuracy obtained from individual features on full images (70.96\%) and far below the peak performance of 77.56\% achieved by LGBM with Feature set 17 on full images. This outcome confirms that even when all descriptors are fused, restricting the analysis to the isolated eye region does not bring the expected benefits. The contextual cues present in surrounding tissues and scales are essential, and their removal through segmentation cannot be fully compensated by feature aggregation.

\begin{table}[H]
\footnotesize
\caption{Accuracy of ML methods using fully combined feature types on segmented images (\%).}
\label{tab:combined_color_features_results_on_segmented_images}
\centering
\begin{tabular}{|l|c|c|c|c|c|c|c|c|}
\hline
\textbf{Feature set} & \textbf{LR} & \textbf{KNN} & \textbf{SVM} & \textbf{ANN} & \textbf{RF} & \textbf{ET} & \textbf{LGBM} & \textbf{CB} \\
\hline 
Feature set 15       &     60.48   &     57.86    &     64.92    &     65.03    &     65.26   &    62.64    &     67.08     &     66.74   \\
\hline
Feature set 16       &     60.59   &     57.63    &     65.15    &     64.92    &     63.44   &    61.96    &     68.11     &     65.95   \\
\hline
Feature set 17       &     59.91   &     58.43    &     64.81    &     66.17    &     64.35   &    61.62    &     66.86     &     64.01   \\
\hline
\end{tabular}
\end{table}

\subsection{Computation time analysis}~\label{sec:computation_time_analysis}

In addition to accuracy, we also analyzed and compared the feature extraction times. This perspective is important because, in practical applications, the trade-off between predictive performance and computational cost often determines the feasibility of deploying a model.

\begin{figure}[H]
   \centering
   \includegraphics[width=1.0\linewidth]{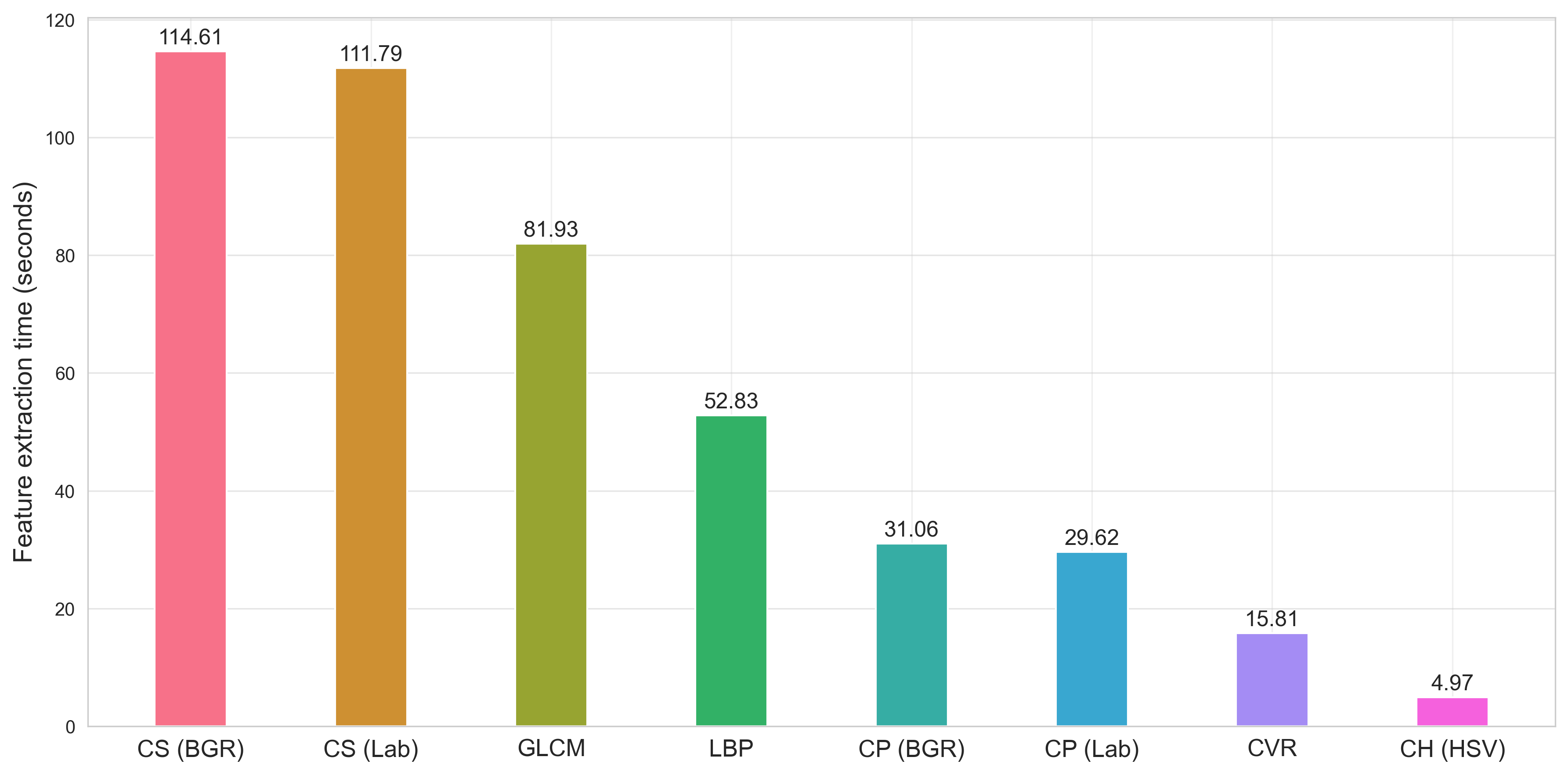}
   \caption{Computation time analysis for the components of the best-performing feature set (FS17).}
   \label{fig:computation_time}
\end{figure}

A detailed examination of Figure~\ref{fig:computation_time} reveals substantial disparities in the computational demands of different feature descriptors. Extracting CS from the BGR and CIELAB spaces is the most time-consuming, taking approximately 114.61 and 111.79 seconds, respectively. Texture-based GLCM is similarly costly, requiring 81.93 seconds due to its pixel-wise calculations of higher-order statistical relationships. By contrast, CH and CVR are much faster, with CH in the HSV space being the most efficient at just 4.97 seconds, and CVR requiring only 15.81 seconds. LBP and CP fall in the middle range, with extraction times between 29 and 53 seconds.

These findings highlight the trade-off between computational cost and classification performance. As shown in Table~\ref{tab:individual_result_on_original_images}, CIELAB CS achieved the highest standalone accuracy (70.96\%) despite its heavy computation. Its inclusion in the final best-performing model (FS17) confirms its critical role in maximizing accuracy. Overall, the computation time analysis not only demonstrates this accuracy-speed trade-off but also provides a practical framework for balancing performance and efficiency in real-world deployment.

This analysis highlights the need to balance accuracy and processing speed when selecting feature descriptors. CS and GLCM offer high accuracy but require significant computation time, while CH and CVR are faster with solid performance. The FS17 model, combining both heavy and lightweight descriptors, achieves optimal accuracy but demands longer processing times.

\subsection{Comparison with previous studies on the same dataset}~\label{sec:comparison_with_previous_studies}

To ensure a fair and faithful comparison with prior work, this section reports results obtained under two different evaluation settings: an augmented data setting combined with cross-validation and a hold-out evaluation on the original, non-augmented dataset, following the experimental protocol adopted in previous studies.

\subsubsection{Re-implementation of prior methodology}

In this subsection, experiments are conducted under the augmented data and cross-validation setting adopted in previous studies (Protocol B).
To ensure a fair and faithful comparison with prior work, we re-implemented the experimental protocol described by Yildiz et al.~\cite{Yildiz2024}, including their strategy of applying mirroring-based data augmentation to the entire dataset before performing 10-fold cross-validation. The hyperparameters of the classifiers inherited from the original study (KNN, LR, SVM, ANN, and RF) were kept identical to those reported in~\cite{Yildiz2024}. For the additional models introduced in this work (ET, LGBM, and CB), appropriate hyperparameter settings were selected. The complete configurations of all classifiers are summarized in Table~\ref{tab:ml_full_image_hyperparameters}.

The results presented in Table \ref{tab:comparison_accuracy} clearly demonstrate the substantial advantage of the proposed handcrafted feature extraction approach over deep feature representations (VGG19, selected as the strongest baseline in~\cite{Yildiz2024}) across multiple ML classifiers. Notably, ANN achieves an accuracy of 97.49\%, outperforming the corresponding VGG19-based model by over 20\%. This significant improvement highlights the effectiveness of the carefully designed feature set, comprising statistical color moments, color histograms, LBP, and GLCM, in capturing the visual cues most relevant to fish eye freshness classification.

Other classifiers further reinforce this trend. For instance, RF attains 91.91\% accuracy, surpassing the VGG19-based counterpart by nearly 24 points, and KNN also benefits from a gain of about 5\%. Even linear models such as LR and SVM perform comparably or slightly better than VGG19 features, demonstrating that the handcrafted features are robust across different learning paradigms. Moreover, ensemble methods like ET, LGBM, and CB reach accuracy levels above 97\%, emphasizing that the proposed features provide discriminative information that can be fully leveraged by tree-based models without requiring deep feature extraction.

\begin{table}[H]
\footnotesize
\caption{Comparison of ML model accuracy (mean $\pm$ standard deviation \%) between the proposed handcrafted features and deep features reported by Yildiz et al.~\cite{Yildiz2024}.}
\centering
\label{tab:comparison_accuracy}
\begin{tabular}{|l|c|c|c|}
\hline
\textbf{Algorithm} & \textbf{Proposed features} & \textbf{VGG19~\cite{Yildiz2024}} & \textbf{Difference} \\ \hline
KNN   & $72.88 \pm 1.48$ & 67.8  & $5.08 \pm 1.48$  \\ \hline
LR    & $63.14 \pm 1.78$ & 65.9  & $-2.76 \pm 1.78$ \\ \hline
SVM   & $73.23 \pm 1.88$ & 74.3  & $-1.07 \pm 1.88$ \\ \hline
ANN   & $\mathbf{97.49 \pm 0.63}$ & 77.3  & $20.19 \pm 0.63$ \\ \hline
RF    & $91.91 \pm 1.16$ & 68.3  & $23.61 \pm 1.16$ \\ \hline
ET    & $\mathbf{97.16 \pm 0.88}$ & -     & -     \\ \hline
LGBM  & $\mathbf{97.78 \pm 0.80}$ & - & -     \\ \hline
CB    & $\mathbf{97.72 \pm 0.74}$ & -     & -     \\ \hline
\end{tabular}
\end{table}

These results can be explained as follows. Under the experimental setting adopted from Protocol B, data augmentation is applied prior to dataset splitting, where image flipping generates duplicated samples originating from the same fish instance and retaining identical class labels. Due to the inherent invariance properties of handcrafted descriptors, such transformations produce identical or near-identical feature vectors.
As a result, the augmented dataset contains redundant samples with identical handcrafted representations appearing across training and test folds during cross-validation. This redundancy is effectively exploited by models with high learning capacity, such as ANN and tree-based ensembles (RF, ET, LGBM, and CB), leading to exceptionally high performance.

In contrast, linear and distance-based classifiers (LR, SVM, and KNN) do not benefit from sample duplication, as their decision boundaries or neighborhood structures remain mathematically invariant when identical samples are repeated. Consequently, the limited or slightly degraded performance of models indicates that the performance gains observed under Protocol B arise from algorithmic exploitation of data redundancy rather than an increase in the intrinsic discriminative power of the handcrafted features.

\subsubsection{Comparative analysis on the non-augmented dataset (Protocol A)}

In addition to the replication experiments conducted under Protocol B, we further evaluated our approach under Protocol A using the standard train-test split proposed by Prasetyo et al.~\cite{Prasetyo2022}, who reported a baseline accuracy of 63.21\% with their custom deep learning architecture, MB-BE (a modified MobileNetV1), on the original, non-augmented dataset.

\begin{figure}[H]
   \centering
   \includegraphics[width=1.0\linewidth]{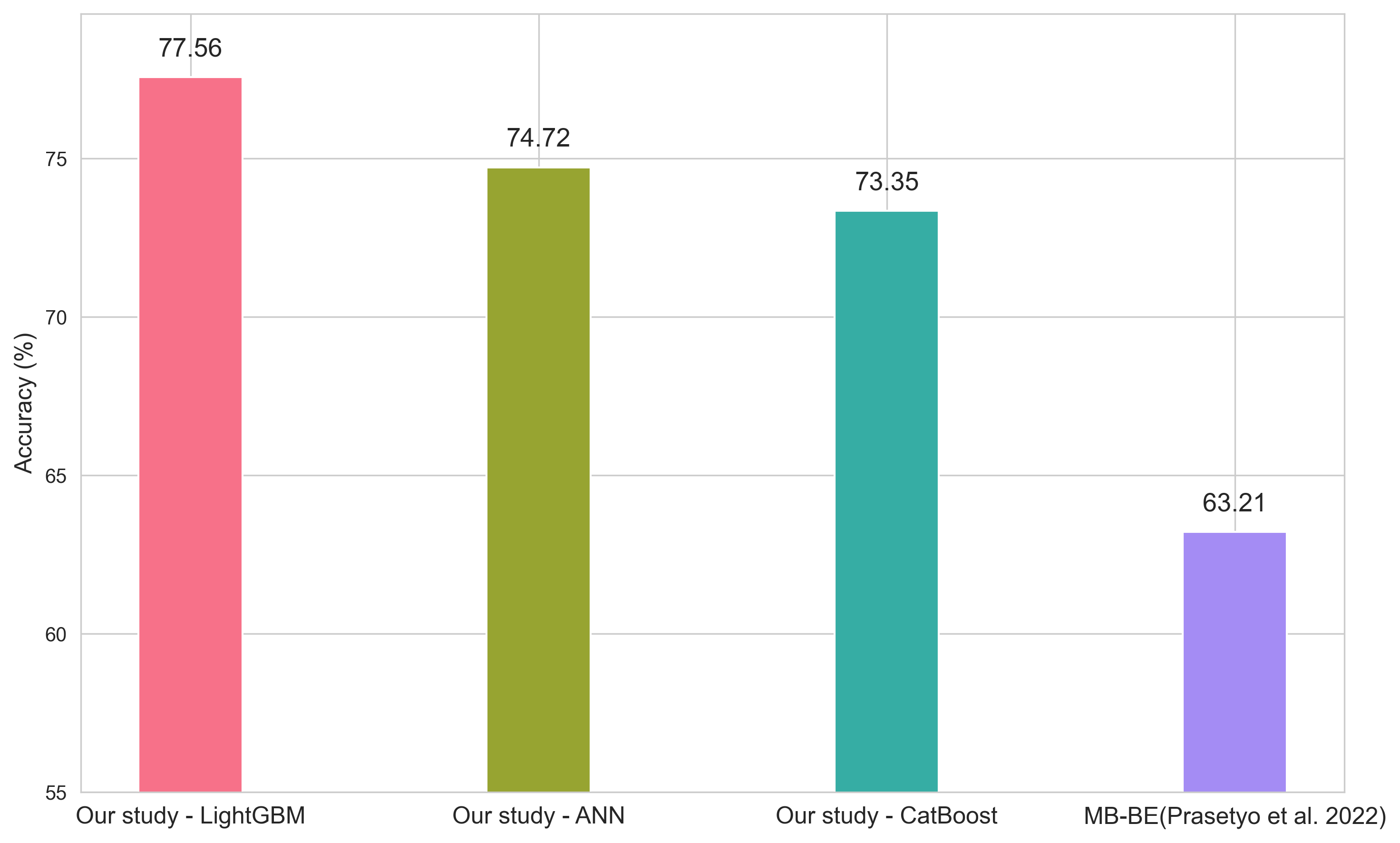}
  \caption{Performance comparison under the experimental setup without data augmentation.}
   \label{fig:compare2}
\end{figure}

The results, summarized in Table~\ref{tab:full_combined_features_results} and visualized in Figure~\ref{fig:compare2}, highlight the remarkable efficacy of our method in this setting as well. Our best-performing model, which combines the handcrafted feature set with a LightGBM classifier, achieved an accuracy of 77.56\%. This marks a substantial improvement of over 14 percentage points compared to the baseline established by Prasetyo et al. Other models in our study, such as ANN (74.72\%) and CatBoost (73.35\%), also comfortably surpassed the previous benchmark.

This outcome is particularly insightful. It demonstrates that a traditional machine learning model, when supplied with a rich and highly discriminative set of handcrafted features, can decisively outperform a complex, custom-designed deep learning architecture. The MB-BE model likely struggled to learn the most salient features from the limited training data. In contrast, our approach explicitly provides this crucial information to the classifier. The superior performance of tree-based ensembles like LightGBM further suggests that they are exceptionally well-suited to model the complex, non-linear relationships inherent in our engineered feature space.

In summary, across two rigorous evaluation settings, our proposed handcrafted feature engineering methodology has established a new state-of-the-art performance on the FFE dataset. The findings collectively underscore that for specialized tasks such as fish freshness assessment, a well-designed, knowledge-driven feature-based method offers a more robust, accurate, and reliable solution.

\section{Conclusion}~\label{sec:conclusion}

This study proposed and systematically evaluated a handcrafted feature engineering approach for fish freshness classification from eye images. We extracted a diverse set of descriptors, including color statistics, variance ratios, percentiles, histograms, and texture features (GLCM and LBP), and integrated them through an incremental fusion strategy to form a rich numerical representation of freshness cues. The extracted features were subsequently used to train and evaluate eight ML models--LR, KNN, SVM, ANN, RF, ET, LGBM, and CB--to comprehensively assess classification performance. 
Experiments on the FFE dataset demonstrated that, under a standard non-augmented train--test setting, the LGBM model achieved an accuracy of 77.56\%, representing an improvement of over 14\% compared to the 63.21\% reported by Prasetyo et al.~\cite{Prasetyo2022}. Under an augmented evaluation protocol consistent with prior work, the ANN model obtained 97.49\% accuracy, compared to the 77.3\% reported by Yildiz et al.~\cite{Yildiz2024}. This augmented performance should be interpreted as an upper-bound estimate, while the non-augmented evaluation provides a more realistic proxy for practical deployment.
Importantly, comparative analyses between full images and ROI-based segmentation revealed that contextual information from surrounding tissues contributes significantly to classification performance, indicating that discriminative freshness cues extend beyond the eye region alone.
Despite the encouraging results obtained under both evaluation protocols, the proposed approach remains laboratory-validated and relies on handcrafted feature engineering, which requires domain expertise and may exhibit limited generalization across unseen species or acquisition conditions. In addition, although handcrafted descriptors demonstrate strong discriminative capability on the FFE dataset, they may fail to capture higher-level abstract patterns that deep learning models can learn automatically when larger and more diverse datasets are available. Future work will therefore focus on hybrid frameworks that combine handcrafted descriptors with deep feature representations, aiming to preserve interpretability while improving robustness and generalization.

\section*{Funding:} This research received no funding.

\section*{Ethical statement:} The authors declare that no ethical approval was required for this study as it exclusively utilizes a publicly available dataset (FFE).

\section*{Data availability:} 

The datasets generated and analyzed during the current study are available in the Mendeley data repository~\citep{MendeleyFFEData}.

\section*{Author contributions:}

    \textbf{Phi-Hung Hoang}: Writing - review \& editing, Writing - original draft, Data curation, Visualization, Validation, Software, Methodology, Investigation.  
      \textbf{Nam-Thuan Trinh}: Writing - review \& editing, Writing - original draft, Data curation, Visualization, Validation, Software, Investigation.  
       \textbf{Van-Manh Tran}: Writing - review \& editing, Writing - original draft, Data curation, Software, Investigation.  
        \textbf{Thi-Thu-Hong Phan}: Conceptualization, Methodology, Supervision, Validation, Writing - original draft, Writing - review \& editing.

\section*{Declaration of Interests:}
 
The authors declare that they have no known competing financial interests or personal relationships that could have appeared to influence the work reported in this paper.

\section*{Declaration of Generative AI and AI-Assisted Technologies in the Writing Process:}

During the preparation of this manuscript, Grammarly and generative large language models were employed to improve grammar and refine wording for clarity. The authors subsequently reviewed and revised the content as necessary and take full responsibility for the final version of the manuscript.

\section*{Additional statement:}
A preprint version of this manuscript will be uploaded to arXiv for early access and citation.

\bibliographystyle{elsarticle-num}
\bibliography{fish-refs}

\end{document}